\documentclass{article}

\PassOptionsToPackage{dvipsnames}{xcolor}

\usepackage{xcolor}

\usepackage[nonatbib,preprint]{neurips_2025}

\usepackage[utf8]{inputenc} 
\usepackage[T1]{fontenc}    
\usepackage{hyperref}       
\usepackage{url}            
\usepackage{booktabs}       
\usepackage{amsfonts}       
\usepackage{nicefrac}       
\usepackage{microtype}      
\usepackage{amsmath}
\usepackage{microtype}
\usepackage{graphicx}
\usepackage{hyperref}
\usepackage{tcolorbox} 
\usepackage{booktabs} 
\usepackage{epsfig}
\usepackage{amssymb}
\usepackage{bm}
\usepackage{amsthm}
\usepackage{multirow}
\usepackage{caption}
\usepackage[super]{nth}

\usepackage{amssymb}%
\usepackage{mathrsfs}%
\usepackage{nicefrac}
\usepackage{color}
\usepackage{pifont}
\usepackage{array}
\usepackage{wrapfig}
\usepackage{mathtools}
\usepackage{colortbl}
\usepackage{mdframed}
\usepackage{makecell}
\usepackage{subcaption}
\usepackage[ruled,noline,nofillcomment]{algorithm2e}
\usepackage{algorithmic}
\usepackage{booktabs}
\usepackage{multirow}

\usepackage{adjustbox}
\usepackage{verbatim}

\definecolor{mycolor_blue}{HTML}{E7EFFA}
\definecolor{mycolor_green}{HTML}{E6F8E0}
\definecolor{mycolor_gray}{HTML}{ECECEC}
\definecolor{pearDark}{HTML}{2980B9}

\title{Flow Diverse and Efficient: \\Learning Momentum Flow Matching via Stochastic Velocity Field Sampling}

\author{%
  Zhiyuan Ma\textsuperscript{1}\footnotemark[1], 
  Ruixun Liu\textsuperscript{2}\footnotemark[1], 
  Sixian Liu\textsuperscript{3},
  Jianjun Li\textsuperscript{4},  
  Bowen Zhou\textsuperscript{1,5}\footnotemark[2] \\
  $^1$ \textnormal{Department of Electronic Engineering, Tsinghua University,} \\
  $^2$ \textnormal{School of Computer Science and Technology, Xi’an Jiaotong University, } \\
  $^3$ \textnormal{Department of Statistics, University of California, Berkeley,} \\
  $^4$  \textnormal{School of Computer Science and Technology, Huazhong University of Science and Technology,} \\
  $^5$  \textnormal{Shanghai Artificial Intelligence Laboratory} \\
   {\tt\small \{mzyth,zhoubowen\}@tsinghua.edu.cn}, \tt\small {liuruixun6343@gmail.com}
}

\begin{document}

\maketitle

\renewcommand{\thefootnote}{\fnsymbol{footnote}}
\footnotetext[1]{Equal Contribution.}
\footnotetext[2]{Corresponding author.}

\begin{abstract}
  Recently, the rectified flow (RF) has emerged as the new state-of-the-art among flow-based diffusion models due to its high efficiency advantage in straight path sampling, especially with the amazing images generated by a series of RF models such as \emph{Flux 1.0} and \emph{SD 3.0}. Although a straight-line connection between the noisy and natural data distributions is intuitive, fast, and easy to optimize, it still inevitably leads to: \textbf{\emph{1) Diversity concerns}}, which arise since straight-line paths only cover a fairly restricted sampling space. \textbf{\emph{2) Multi-scale noise modeling concerns}}, since the straight line flow only needs to optimize the constant velocity field $\bm v$ between the two distributions $\bm\pi_0$ and $\bm\pi_1$. In this work, we present Discretized-RF, a new family of rectified flow (also called momentum flow models since they refer to the previous velocity component and the random velocity component in each diffusion step), which discretizes the straight path into a series of variable velocity field sub-paths (namely \emph{``momentum fields''}) to expand the search space, especially when close to the distribution $p_\text{noise}$. Different from the previous case where noise is directly superimposed on $\bm x$, we introduce noise on the velocity $\bm v$ of the sub-path to change its direction in order to improve the diversity and multi-scale noise modeling abilities. Experimental results on several representative datasets demonstrate that learning momentum flow matching by sampling random velocity fields will produce trajectories that are both diverse and efficient, and can consistently generate high-quality and diverse results. Code is available at \url{https://github.com/liuruixun/momentum-fm}.

\end{abstract}
\section{Introduction}
Flow-based diffusion models~\cite{lipmanflow,bartosh2024neural,luo2024flowdiffuser,liu2023instaflow} have recently attracted widespread attention, which generate a wide variety of realistic natural images from pure noise distributions by modeling the trajectory from noise distributions to data distributions. As a milestone work, the most popular flow models currently are the rectified flow (RF) models~\cite{liu2023flow} built upon straight-line trajectories, which significantly improves the sampling efficiency by establishing the shortest straight-line connection between the noise distribution $\bm\pi_0$ and the data distribution $\bm\pi_1$, and the model can be easily optimized by simply calibrating this straight-line trajectory $d\bm x_t/dt=\bm v_\theta$ at a constant rate $\bm x_1-\bm x_0$. Due to its high sampling efficiency (even enabling one-step diffusion generation), RF is also considered one of the fastest flow-based optimal transport models. 

Though remarkable success has been witnessed, RF still suffers from limitations in \emph{diversity} and \emph{multi-scale noise modeling}. Specifically, \textbf{\emph{1) Diversity concerns}}, which arise since straight-line paths only cover a fairly restricted sampling space. \textbf{\emph{2) Multi-scale noise modeling concerns}}, since the   
\begin{wrapfigure}{r}{0.5\textwidth} 
  \centering
  \includegraphics[width=\linewidth]{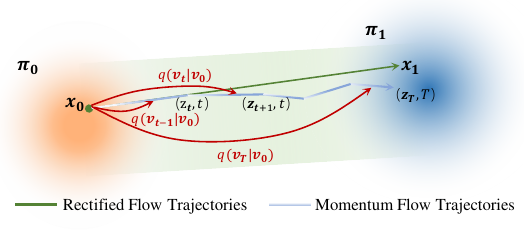}
  \vspace{-10pt} 
  \caption{Graphical momentum flow trajectories. \\ 
  \footnotesize \textbf{\textcolor{RoyalBlue}{Momentum Flow (blue)}} \emph{vs.} \textbf{\textcolor{ForestGreen}{Rectified Flow (green)}}.}
  \label{fig:wrap}
  \vspace{-6pt}
\end{wrapfigure}
straight flow only needs to directly optimize the constant velocity field $\bm v_\theta\rightarrow (\bm x_1-\bm x_0)$ between the two distributions $\bm\pi_0$ and $\bm\pi_1$, instead of considering multi-scale progressive denoising. At the other extreme, the diffusion probability models (\emph{e.g.,} DDPM) based on fluctuation trajectories have extremely strong diversity and multi-scale noise modeling capabilities but face the problem of training- and sampling-efficiency because they require a large number of time steps to sample, and each step should be optimized to achieve high-fidelity modeling of the reverse trajectory.
In this work, in order to strike a balance and take into account both efficiency and diversity (especially the potential diversity when close to noise distribution $\bm\pi_1$), we propose a Discretized-RF model, also known as the momentum flow matching model. For clarity, we first give a unified definition of the flow transport problem and then introduce our momentum flow transport.

\textbf{Flow Transport Problem Definition.} \emph{Given empirical observations of two distributions $\bm x_0\sim\bm\pi_0$ (real data distribution) and $\bm x_1\sim\bm\pi_1$ (noise distribution) on $\mathbb{R}^d$, find a transport trajectory $T_o: \mathbb{R}^d\rightarrow \mathbb{R}^d$ that satisfies $\bm x_1:=T(\bm x_0)\sim\bm\pi_1$ when $\bm x_0\sim\bm\pi_0$. At the same time, the flow transport should own the estimable property of the reverse solution trajectory, that is, $\bm x_0:=\tilde{T}_\theta(\bm x_1)\sim\bm\pi_0$ when $\bm x_1\sim\bm\pi_1$, which requires the trajectory to be continuous and tractable.
}

\textbf{Momentum Flow Transport (Discritized-RF).} \emph{
Given the shortest optimal transport $d\bm x_t/dt=\bm v$ (straight-line trajectory) at a constant rate $\bm v=\bm x_1-\bm x_0$ and a series of discretized anchor points $\{\bm z_1,\cdots, \bm z_{T-1}\}$, find a segmented straight-line trajectory $T_{\bm x_0\mapsto \bm x_1}=\{\bm x_0,\bm z_1,\cdots,\bm z_{T-1},\bm x_1\}$ that satisfies $d\bm z_t/dt=\bm v_t$, $\bm v_t=\sqrt{\gamma} \bm v_{t-1}+\sqrt{(1-\gamma)}\bm\epsilon_t, \bm\epsilon_t\sim\mathcal{N}(0,\bm I)$. Meanwhile, the endpoint transport of this momentum flow trajectory are respectively defined as: $T_{\bm x_0\mapsto \bm z_1}:d\bm z_t/dt=\bm v_0$ (initialized by $\bm x_1-\bm x_0$) and $T_{\bm z_{T-1}\mapsto \bm x_1}:d\bm z_t/dt=\bm \epsilon, \bm\epsilon\sim\mathcal{N}(0,\bm I)$. The momentum flow ensures that the velocity is Gaussian divergent when approaching $\bm\pi_1$, while the velocity is more deterministic and faster when approaching $\bm\pi_0$. Note the acceleration $\bm \epsilon$ follows the same Gaussian distribution $\mathcal{N}(0,\bm I)$ and can therefore be easily estimated by the neural model $\bm\epsilon_\theta$ to obtain a tractable inverse trajectory $\tilde{T}_{\bm x_1\mapsto \bm x_0;\theta}=\{\bm x_1,\bm z_{T-1;\theta},\cdots,\bm z_{1;\theta},\bm x_{0;\theta}\}$.
}

The goal of this work is to extend the constant velocity field model to the acceleration field model by learning the momentum flow matching via stochastic velocity field sampling, so as to finally derive an optimal transport path with both speed and diversity. The main contributions are summarized below:
\begin{itemize}
    \item \textbf{The trade-off between efficiency and diversity:} Is the straighter the flow, the better? Although the straight line between two points is the shortest in the rectified flow optimization, and it is obviously easier to optimize this straight line, it cannot be ignored that it still suffers from the low diversity of sampling (the trajectory variance is close to 0, that is, the constant velocity field) and the limited quality caused by the deterministic trajectory.
    \item \textbf{The optimal approximation of multi-scale noise between a straight line and a fluctuating line.} The discretized rectified flow (Discretized-RF) solution trajectory is a better approximation of the multi-scale noise-adding (denoising). It is easier to optimize than the stochastic differential equation (i.e., \textit{fluctuation flow trajectory}) and can better model multi-scale noise than the constant velocity field differential equation (i.e., \textit{rectified flow trajectory}). Therefore, solving an optimal discretized rectified flow (segmented trajectory, each small segment is a straight line) can better model multi-scale noise and is easy to solve.
    \item \textbf{Variable momentum flow matching and constant acceleration field prediction}: The forward discretized rectified flow can be easily constructed by sampling the random velocity component and the previous velocity component (i.e., momentum), and the reverse momentum flow trajectory can be easily sampled by our momentum flow matching algorithm.
    \item \textbf{Superior performance on multiple image datasets}: Empirically, the Momentum Flow model achieves competitive FID and recall scores with substantially fewer denoising steps. On multiple image datasets, including CIFAR-10~\cite{krizhevsky2009learning} and CelebA-HQ~\cite{karras2018progressive}, it consistently matches or even outperforms the performance of Rectified Flow while requiring only half the number of sampling steps.
\end{itemize}

\section{Related Work}

\textbf{Diffusion Models: High Diversity at the Cost of Efficiency.} 
Diffusion models~\cite{song2019generative,ho2020denoising,song2020score,nichol2021improved,kawar2022denoising,ma2024neural,ma2024safe,ma2025efficient,ma2024adapedit} have emerged as a powerful class of generative models, known for their impressive sample diversity. However, their stochastic diffusion trajectories typically require hundreds or thousands of sampling steps, leading to significant computational costs. To overcome this inefficiency, researchers have proposed some optimization methods along two primary directions: sampling acceleration strategies~\cite{liu2022pseudonumericalmethodsdiffusion,salimans2022progressivedistillationfastsampling,NEURIPS2023_d6f764aa,Meng_2023_CVPR,song2023consistency,sauer2024adversarial,xu2024ufogen} and model architecture improvements~\cite{li2023snapfusion,zhao2024mobilediffusion,xu2024ufogen,li2024distrifusion,ma2024lmd}. For instance, DDIM~\cite{song2020denoising} introduces a non-Markovian reverse process that decouples temporal dependencies, substantially reducing the number of sampling steps. DeepCache~\cite{Ma_2024_CVPR} accelerates inference by caching and retrieving features across adjacent denoising stages to avoid redundant computations. On the architectural side, some works enhance model efficiency by employing custom multi-decoder U-Net designs that combine time-specific decoders with a shared encoder~\cite{Zhang_2024_CVPR}, or by enabling parallel decoder execution to speed up the denoising process~\cite{NEURIPS2024_9ad996b5}. Despite these advances, diffusion-based models still rely on curved stochastic paths, which remain inherently more expensive to compute than deterministic or straight-path methods. As a result, the fundamental trade-off remains: high sample diversity comes at the expense of computational efficiency.

\textbf{Rectified Flows: Faster Sampling Meets Less Diversity.} 
Rectified flow models~\cite{liu2022rectifiedflowmarginalpreserving,liu2023flow,liu2023instaflow,wang2024rectifieddiffusionstraightnessneed,Zhu_2024_CVPR,NEURIPS2024_f0d629a7} significantly improve sampling efficiency over traditional diffusion models by optimizing straight-line trajectories in probability space. However, their deterministic and straight sampling paths fundamentally limit their sampling diversity. To address this limitation, various techniques have been proposed to enhance sample diversity while maintaining efficiency. Some methods focus on optimizing noise sampling techniques~\cite{yan2024perflow,wang2024frieren,liu2024rfwavemultibandrectifiedflow}. For example, training on perceptually relevant noise scales~\cite{esser2024scaling}, or sampling from multi-modal flow directions~\cite{guo2025variationalrectifiedflowmatching}. Other efforts aimed at improving generation quality~\cite{lee2024improving,li2024omniflow,dalva2024fluxspacedisentangledsemanticediting} include applying flow matching in the latent space of pretrained autoencoders~\cite{dao2023flowmatchinglatentspace}, mitigating numerical errors in the ODE-solving process~\cite{wang2024tamingrectifiedflowinversion}, or introducing posterior-mean-based optimal estimators~\cite{ohayon2025posteriormeanrectifiedflowminimum}. However, the trade-off between sampling speed and diversity persists, motivating the development of adaptive flow-based methods that preserve computational efficiency while enhancing sampling diversity. 

Unlike previous methods, our momentum flow matching model introduces a momentum field into the forward process, where multi-scale noise dynamically adjusts the trajectory directions to promote sampling diversity. To improve computational efficiency, the reverse trajectory is discretized into multiple sub-paths, each optimized via rectified flow. As a result, our model retains the fast sampling speed of rectified flow while recovering much of the sample diversity achieved by diffusion models.

\section{Method}
In this section, we propose Momentum Flow Transport, a novel flow-based diffusion model family that aims to achieve an optimal balance between diversity and efficiency via a brand-new momentum flow matching technique in~Sec.~\ref{method:momentum}. Momentum Flow is a dynamically optimal approximation of multi-scale noise-adding (or de-noising) between a straight line and a fluctuating line by combinating: 1) \emph{fluctuation flow trajectory \textbf{(close to $\bm x_1$)}} for diversity and 2) \emph{rectified flow trajectory \textbf{(close to $\bm x_0$)}} for efficiency. We then further introduce the momentum-guided forward process in Sec.~\ref{method:forward_process} and the acceleration fields-driven reverse process in Sec.~\ref{method:reverse_process}. We provide a detailed discussion below.

\subsection{Momentum Flow Matching}
\label{method:momentum}

\textbf{Optimal Transport (OT).}
The optimization problem from the noise distribution $\bm \pi_1$ to the data distribution $\bm \pi_0$ can be regarded as an optimal transport (OT) problem. Since it is extremely difficult to directly solve the trajectory from $\bm \pi_1$ to $\bm \pi_0$, recent flow-based methods~\cite{lipmanflow,liu2023flow} usually first give a tractable forward trajectory $T_{o}$ to transport any $\bm x_0\sim\bm\pi_0$ to $\bm x_1\sim \mathcal{N}(0,\bm I)$ (approximation of $\bm\pi_1$), and then solve the posterior $p(\bm\pi_0|\bm\pi_1)=\tilde{T}_\theta(\bm\pi_1)$ via a flow-matching trajectory estimator $\tilde{T}_\theta$, 
\begin{equation}
\bm\pi_1=T_o(\bm\pi_0)=\int d\bm z_t T_o\left(\bm \pi_1 \mid \bm z_t\right) \bm \pi\left(\bm z_t\right),
    \label{eq:OT}
\end{equation}
\begin{equation}
    \bm\pi_0=\tilde{T}_\theta(\bm\pi_1)=\int d \bm z_{(0: T)}\bm \pi\left(\bm z_T\right) \prod_{t=1}^T p\left(\bm z_{t-1} \mid \bm z_t\right).
\end{equation}
\textbf{Stochastic Transport (\emph{Diversity}-OT) and Rectified Transport (\emph{Efficiency}-OT).} 
Stochastic Transport~\cite{ho2020denoising,song2020denoising} and Rectified Transport~\cite{liu2023flow,liu2023instaflow} are two common optimal transport methods, which are respectively known for their high sampling quality (\emph{diversity}) and fast sampling speed (\emph{efficiency}). However, they all struggle with the balance between efficiency and diversity, either relying on overly divergent sampling steps (trajectory variance $\bm \beta_T\to \infty$) or relying on predefined straight trajectories (trajectory variance $\bm \beta_T=0$). The core of this work is to find an optimal trajectory $T_{o}$ in terms of efficiency and diversity so that the trajectory variance tends to $0$ when it approaches data distribution $\bm\pi_0$ (for \emph{efficiency}) and tends to $\infty$ when it approaches noise distribution $\bm\pi_1$ (for \emph{diversity}).
\begin{figure}[t]
    \centering
    \includegraphics[width=1\textwidth]{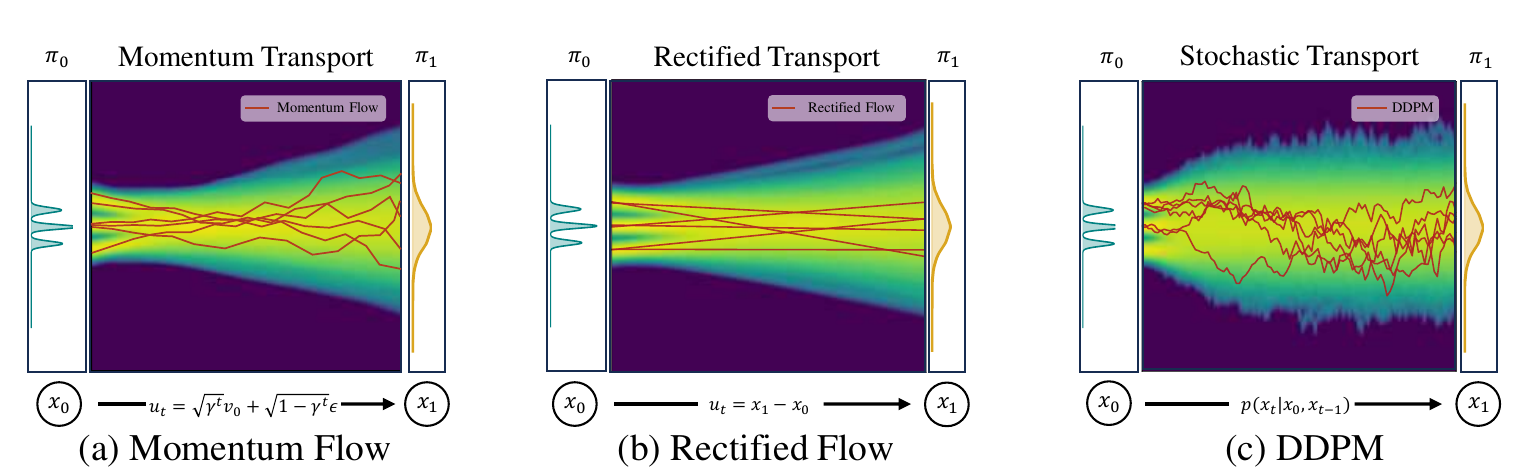}
    \vspace{-16pt}
    \caption{
    Overview of Momentum Flow. Compared with Rectified Flow~\cite{liu2023flow} (\emph{Efficiency}-OT) and DDPM~\cite{ho2020denoising} (\emph{Diversity}-OT), the momentum flow tends to explore diversity when it is close to noise distribution $\bm \pi_1$, and tends to focus on efficiency when it is close to data distribution $\bm\pi_0$.
    } 
    \label{fig:method}
\vspace{-16pt}
\end{figure}

\textbf{Momentum Field (Acceleration Field).} 
In order to find a balanced optimal trajectory, we introduce the momentum field. That is a variable velocity field referring to the previous velocity component and the random velocity component in each diffusion step. Let $\bm \nu=\{\bm v_t\}_0^{T-1}$ represent the momentum field (for guiding $\bm x_0$ to $\bm x_1$), $\bm v_t$ denote the velocity vector from time $t$ to time $t+1$, we have:
\begin{equation}
\frac{d\bm z_t}{dt}=\bm v_t,\;\;
\bm v_t=
\begin{cases}
    \beta(\bm\epsilon_0-\bm x_0) & \text{if } t=0\\
    \sqrt{\gamma_t} \bm v_{t-1}+\sqrt{1-\gamma_t}\beta\bm\epsilon_t & \text{if } 0< t < T\\
    \beta\bm\epsilon_T & \text{if } t=T.
    \label{eq:dzdt}
    \end{cases}
\end{equation}
Here $\bm z_t\sim\bm\pi(\bm z_t)$ is the middle noise-perturbed distribution during the forward diffusion process, and $\left\{\gamma_t\right\}_{1}^{T-1}$ is the momentum decay coefficient, which can be chosen as a constant $\gamma$ smaller than $1$ or a positive decreasing series. We choose the constant $\gamma$ in our work. For convenience, $\beta$ denotes the normalization coefficient $\beta:=(\sqrt{\gamma}-1)/(\sqrt{\gamma^T}-1)$, and $\{\bm\epsilon_t\}_0^T\sim\mathcal{N}(0,\bm I)$ denotes the standard Gaussian noises. Under the influence of this momentum field, for $\forall \bm x_0\sim\bm\pi_0,\bm x_1\sim\bm\pi_1$, the data $\bm x_0$ will gradually transform into noise $\bm x_1$ through the trajectory ${T}_{\bm x_0\mapsto \bm x_1}=\{\bm x_0,\bm z_1,\cdots,\bm z_{T-1},\bm x_1\}$. Note this momentum field $\left\{\bm v_t\right\}_{0}^{T-1}$ maintains the dynamics of the rectified flow~\cite{liu2023flow} during the initial noise-adding stage with the fastest initial vector $\bm v_0=\beta(\bm\epsilon_0-\bm x_0)$. As the velocity $\bm v_0$ is gradually noise-perturbed until it approaches a Gaussian noise $\beta \bm\epsilon_T$, we complete the progressively diverse modeling of an OT trajectory ${T}_{\bm x_0\mapsto \bm x_1}$. 
Similar to DDPM~\cite{ho2020denoising}, we can directly obtain the momentum $\bm v_t$ at any timestep $t$ via the one-step update formula as (see Appendix~\ref{appendix:A} for details),
\begin{equation}
    \bm v_t = \sqrt{\bar{\gamma_t}} \bm v_0+\sqrt{1-\bar{\gamma_t}}\beta\bm\epsilon_t,\;\; \bm v_0 = \beta(\bm \epsilon_0 - \bm x_0),
    \label{eq:3}
\end{equation}
where $\bar{\gamma_t} :=\prod_{i=1}^t \gamma_i$. As derived from Eq.~\eqref{eq:3}, the proportion of $\bm v_0$ in $\bm v_t$ decays exponentially with increasing $t$. This indicates that during the forward process, the momentum $\bm v_t$ gradually deviates from the linear direction defined by $\bm v_0$, thereby progressively expanding the exploration diversity.

\textbf{Momentum Flow Matching Objective.}
Building upon the flow matching framework for velocity field regression, we optimize the optimal transport (OT) problem by minimizing the MSE between predicted momentum and ground-truth. The momentum flow matching objective is formulated as: 
\begin{equation}
    \mathcal{L}_{\text{MFM}}(\theta) = \mathbb{E}_{t \sim U[0,1]} \| \bm{u}_{\theta}(\bm{z}_t, t) - \bm v_t \|^2,
    \label{eq:loss}
\end{equation}
where $\theta$ denotes learnable parameters for neural network $\bm{u}_\theta(\cdot,t)$, and $t\sim\mathcal{U}[0,1]$. In the inference phase, once the momentum estimate $\bm v_{\theta;t}=\bm{u}_\theta(\cdot,t)$ is obtained, the reverse OT trajectory $\tilde{T}_\theta$ can be derived, as detailed in the subsequent forward and reverse processes.
\subsection{Forward Process of the Momentum Flow}\label{method:forward_process}
Let $\bm z_0=\bm x_0\sim\bm\pi_0$  and $\bm z_T =\bm x_1\sim\bm\pi_1$ respectively denote the data and noise distributions on $\mathbb{R}^d$. When applying our momentum flow to the forward diffusion process, we can obtain the intermediate noisy distribution $\left\{\bm z_t\sim\bm\pi({\bm z_t})\right\}_{1}^{T-1}$ at discretized anchor points $\bm z_t$. In general, the forward diffusion process is defined as a Markov chain that progressively injects Gaussian noise $\bm \epsilon$ into $\bm x_0$ over $T$ timesteps according to the forward coefficient $a_t$ and $b_t$:
\begin{equation}
\setlength\abovedisplayskip{-2pt}
    q(\bm z_t \vert \bm z_{t-1}) = \mathcal{N}(\bm z_t; a_t \bm z_{t-1}, b_t^2\mathbf{I}), \quad \quad q(\bm z_{(1: T)}|\bm z_0)= \prod_{t=1}^T q(\bm z_t|\bm z_{t-1}).
\end{equation}
We subsequently introduce our momentum field into the classical diffusion process to adjust the balance between diversity and efficiency for the exploration of the forward optimal trajectory $T_o$, the detailed steps are presented in  Algorithm~\ref{alg:FP}. 

\textbf{Forward Momentum Flow.}
From Eq.~\eqref{eq:3}, we can observe that the recursive formulation of our momentum field shares the similar form as that in DDPM~\cite{ho2020denoising}, allowing us to directly obtain the prior probability distribution $q(\bm v_t \vert \bm v_{t-1})$ of the momentum flow:
\begin{equation}
    q(\bm v_t \vert \bm v_{t-1}) = \mathcal{N}(\bm v_t; \sqrt{\alpha_t} \bm v_{t-1}, (1-\alpha_t)\beta^2\mathbf{I}), \quad \quad q(\bm v_t \vert \bm v_{0}) = \mathcal{N}(\bm v_t; \sqrt{\bar{\alpha_t}} \bm v_{0}, (1-\bar{\alpha_t})\beta^2\mathbf{I}),
\end{equation}
where $\alpha_t := \gamma$ and $\bar{\alpha_t} :=\prod_{i=1}^t \alpha_i = \gamma ^t $ ($\gamma<1$ is a fixed constant). Notably, the formal alignment between the momentum flow and the forward process in DDPM~\cite{ho2020denoising} also allows a straightforward derivation of the posterior distribution $p_\theta(\bm v_{t-1} \vert \bm v_t)$ of the momentum flow, refer to Sec.~\ref{part1}.

\begin{algorithm}[t]
\caption{Momentum Flow Transport: Forward Process}\label{alg:FP}
\begin{algorithmic}[1]
\STATE \textbf{Procedure}: ${T}_o = \mathtt{MomentumField}((\bm z_0,\bm z_T))$: 
\STATE \textbf{Input: } 
$\bm z_0 \sim \bm \pi_0$, $\bm z_T = \bm \epsilon_0 \sim \bm \pi_1$, $T$, $\{\bm \gamma_t\}_1^{T-1}$, $\beta$, $\bm v_0 = \beta(\bm \epsilon_0 - \bm z_0)$.
\STATE \textbf{For $t \leftarrow 1$ to $T$ do repeat noise disturbance:}
    \begin{itemize}
        \item $\bm \epsilon_t \sim \mathcal{N}(0,\bm I)$.
        \item $\bm z_t = \bm z_{t-1} + \bm v_{t-1}$.
        \item $\bm v_t = \sqrt{\gamma_t}\bm v_{t-1} + \sqrt{1-\gamma_t}\beta\bm\epsilon_t $.
    \end{itemize}
\STATE \textbf{Return:} Trajectory ${T}_o=\{\bm z_0,\bm z_1,\cdots,\bm z_{T-1},\bm z_T\}$.
\end{algorithmic}
\end{algorithm}

\textbf{Forward Data Flow.}
Based on the above momentum flow, we can further build the forward trajectory (i.e., \emph{data flow}) ${T}_{o}=\{\bm z_0,\bm z_1,\cdots,\bm z_{T-1},\bm z_T\}$, which is represented in the form of a conditional probability distribution $q(\bm z_t \vert \bm z_{t-1})$. According to Eq.~\eqref{eq:dzdt} and Eq.~\eqref{eq:3}, the one-step forward data distribution can be obtained (see Appendix ~\ref{appendix:B} for a derivation) as,
\begin{equation}
    q(\bm z_t \vert \bm z_{0}) = \mathcal{N}\left(\bm z_t; (1-(\frac{\sqrt{\gamma^t}-1}{\sqrt{\gamma}-1}) \beta) \bm z_{0}, ((\frac{\sqrt{\gamma^t}-1}{\sqrt{\gamma}-1})^2-\frac{\gamma^t-1}{\gamma-1}+t)\beta^2\mathbf{I}\right).
    \label{eq:oneward}
\end{equation}
 Due to $\beta := (\sqrt{\gamma}-1)/(\sqrt{\gamma^T}-1)$, when computing the noise distribution $\bm \pi(\bm z_T) $, we can eliminate the complex coefficient in front of $\bm{z}_0$ in Eq.~\eqref{eq:oneward} to derive a \emph{zero}-mean Gaussian distribution (independent of the data distribution $\bm \pi_0$):
\begin{equation}
    q(\bm z_T \vert \bm z_0)=\mathcal{N}\left( \bm z_T;0,((\frac{\sqrt{\gamma^T}-1}{\sqrt{\gamma}-1})^2-\frac{\gamma^T-1}{\gamma-1}+T)\beta^2\mathbf{I}\right),
\end{equation}
This simplified formula facilitates the subsequent reverse momentum transport process and significantly reduces computational complexity during training and inference.

\subsection{Reverse Process of the Momentum Flow}\label{method:reverse_process}
\label{part1}
The reverse process of the momentum flow aims to restore noise distribution $\bm \pi_1$ to data distribution $\bm \pi_0$ via an inverse trajectory $\tilde{T}_{\theta}=\{\bm z_T,\bm z_{T-1;\theta},\cdots,\bm z_{1;\theta},\bm z_{0;\theta}\}$, which is estimated by a neural network for approximating $\bm z_{t;\theta} \sim \bm\pi(\bm z_{t;\theta})$. To achieve this, we can approximate the momentum field $\{\bm v_t\}_0^{T-1}$ and then utilize the relationship $\bm z_{t-1;\theta} = \bm z_{t;\theta} -  \bm v_{t-1;\theta}$ to estimate $\bm z_{t-1;\theta}$ from $\bm z_{t;\theta}$ and $\bm v_{t-1;\theta}$, as illustrated in Algorithm~\ref{alg:MFM}. We denote the estimated values of $(\bm z_t, \bm v_t)$ as $(\bm z_{t;\theta},\bm v_{t;\theta})$. Based on this framework, we discuss two ways to approximate the momentum field. The first way is to approximate $\bm v_t$ by estimating $p_\theta(\bm v_{t-1} \vert \bm v_t)$. Benefiting from the formal similarity between the forward momentum flow and the DDPM formulation~\cite{ho2020denoising}, we can derive directly the corresponding posterior distribution and estimate $\bm v_{t-1;\theta}$ from $\bm v_{t;\theta}$ by training a noise predictor $\bm \epsilon_\theta$:
\begin{equation}
    p_\theta(\bm v_{t-1} \vert \bm v_t) = \mathcal{N}\left(\frac{\sqrt{\gamma}\left(1-\gamma^{t-1}\right) + \sqrt{\gamma^{t-1}}(1-\gamma)}{1-\gamma^t}\bm v_t- \frac{(1-\gamma)\beta \bm \epsilon_\theta}{\sqrt{\gamma(1-\gamma^t)}},\frac{(1-\gamma)\left(1-\gamma^{t-1}\right)}{1-\gamma^t}\beta^2\mathbf{I}\right),
\end{equation}
\begin{equation}
    \bm v_{t-1;\theta} = \frac{\sqrt{\gamma}\left(1-\gamma^{t-1}\right) + \sqrt{\gamma^{t-1}}(1-\gamma)}{1-\gamma^t}\bm v_{t;\theta}- \frac{(1-\gamma)\beta \bm \epsilon_\theta}{\sqrt{\gamma(1-\gamma^t)}}+\sqrt{\frac{(1-\gamma)\left(1-\gamma^{t-1}\right)}{1-\gamma^t}}\beta\bm \epsilon.
\end{equation}
The second method directly approximates $\bm v_t$ by employing rectified flow on each sub-path. Specifically, between each adjacent intermediate noise-perturbed distribution pair $(\bm\pi({\bm z_t}),\bm\pi(\bm z_{t-1}))$ at the discretized anchor point pair $(\bm z_t,\bm z_{t-1})$, we insert $M$ intermediate points $\bm z_{t-1}^{(m)}$ via linear interpolation:
\begin{equation}
    \bm z_{t-1}^{(m)} = m\bm z_t + (1-m)\bm z_{t-1},
\end{equation}
where $m \sim \mathcal{U}[0,1]$. To enhance sampling efficiency, we apply rectified flow to formulate a straight path for each sub-path $\{{\bm\pi({\bm z_t}) \to \bm\pi({\bm z_{t-1}})}\}_1^{T}$, with the network $\bm u_{\theta}(\cdot,t)$ trained to match the corresponding velocity $\bm v_{t-1} = \bm z_t - \bm z_{t-1}$. Therefore, the original objective~\eqref{eq:loss} is reformulated into the following optimization objective:
\begin{equation}
    \mathcal{L}_{\text{MFM}}(\theta) =
    \sum_{t=1}^T\mathbb{E}_{t \sim \mathcal{U}[0,1]}\left[ \|\bm u_{\theta}(m\bm z_{t} + (1-m)\bm z_{t-1}, m) -(\bm z_{t} - \bm z_{t-1}) \|^2 \right].
\end{equation}
The second method achieves much higher computational efficiency than DDPM by using rectified flow to optimize the trajectory. Therefore, we follow the second method in all experiments. 

\begin{algorithm}[t]
\caption{Momentum Flow Matching: Reverse Process}\label{alg:MFM}
\begin{algorithmic}[1]

\STATE \textbf{Procedure}: $\tilde{T}_\theta = \mathtt{MomentumFlow}((\bm z_0,\bm z_T))$: 
\STATE \textbf{Input: } 
Momentum model $\bm u_{\theta}(\cdot,t): \mathbb{R}^d \times [0,1] \to \mathbb{R}^d$ with parameters $\theta$.
\STATE \textbf{Training:}
        $\hat{\theta} = \arg\min_\theta \sum_{t=1}^T\mathbb{E}\left[ \|\bm u_{\theta}(m\bm z_{t} + (1-m)\bm z_{t-1}, m) -(\bm z_{t} - \bm z_{t-1})\|^2 \right]$, 
        \\with $m \sim \mathcal{U}[0,1]$.
\STATE \textbf{For $t \leftarrow T$ to $1$ do repeat sampling:}
    \begin{itemize}
        \item Draw $(\bm z_{t-1}, \bm z_t)$ from $\bm \pi(\bm z_{t-1}) \times \bm \pi(\bm z_{t})$, with $\bm z_{t-1}\sim \bm \pi(\bm z_{t-1})$ and $\bm z_{t}\sim \bm \pi(\bm z_{t})$.
        \item Solve ODE: $\frac{d\bm z_t}{dt} = \bm u_\theta(\bm z_t^m, m)$, with $\bm z_0 \sim \bm \pi_0$.
        \item Return: Sub-trajectory $\bm z_t = \{\bm z_t^m : m \in [0, 1]\}$.
    \end{itemize}
\STATE \textbf{Return:} Trajectory $\tilde{T}_\theta = \{\bm z_t : t \in [0, 1]\}$.
\end{algorithmic}
\end{algorithm}
\section{Experiments}
\label{exp}
In this section, we conduct experiments using the rectified flow framework implemented in PyTorch to evaluate the image generation diversity and efficiency of the proposed momentum flow model. The primary objectives are to compare the generating performance between momentum flow and rectified flow, and to analyze the impact of the momentum field on the diversity and speed of the generative process. The results show that momentum flow retains the fast sampling capability of straight velocity fields. In addition, by injecting multi-scale noise through the momentum fields, the diversity and the quality of the generated images are significantly enhanced.

\subsection{Toy Example}
\label{exp:1}
\begin{figure}[t]
    \centering
    \includegraphics[width=1\textwidth]{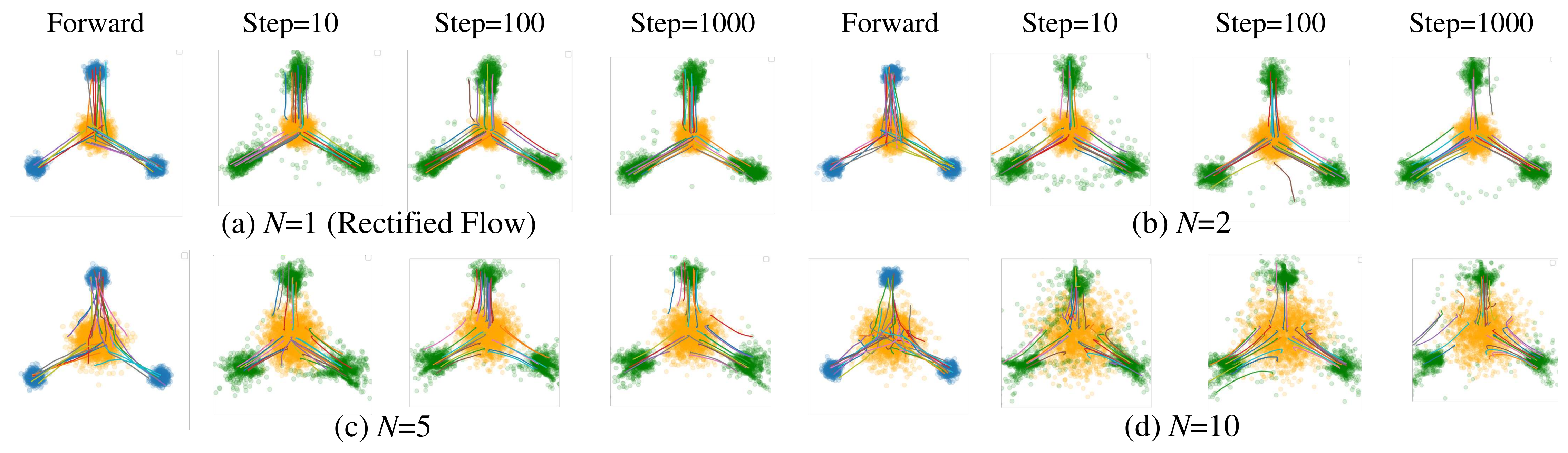}
    \vspace{-16pt}
    \caption{Forward and  reverse trajectories of momentum flow with different numbers $N$ of discretized anchor points. Blue points sample from $\bm \pi_0$, orange points sample from $\bm \pi_1$, green points denote generated samples, and lines represent transport trajectories.} 
    \label{fig:ex1}
\vspace{-16pt}
\end{figure}

To illustrate the theoretical background of momentum flow, we provide an example in Figure~\ref{fig:ex1}, showing the expected momentum flow in the forward process and the optimal transport path in the reverse process. Momentum flow is simulated using the Euler method with a constant step size of $1/N$, computed at $N$ discrete anchor points, where $N$ denotes the number of such points, corresponding to the value of $T$ in Momentum Flow. The number of sampling points on each segment is defined as $\hat{Step}$, and the total number of sampling steps is $\hat{Step} \times N = Step$. In all experiments in this section, this notation is used by default. In this section, we define the use of a fully connected neural network with two hidden layers to estimate the momentum field. The model is trained using full-batch gradient descent and optimized with the Adam optimizer.

As shown in Figure~\ref{fig:ex1}, increasing the number of discretized anchor points causes significant fluctuations in the velocity field near 
$\bm \pi_0$ during the forward process, highlighting the impact of trajectory complexity on learning. Furthermore, our method encourages exploration of diverse trajectories, as evidenced by the "turning-back" phenomenon observed in the early stages of the reverse process when the number of discretized anchor points ($N$) increases. This allows for more exploration in the space rather than directly pointing to the  $\bm \pi_0$ distribution. Despite the unpredictability of the varying velocity field, the residual correlation between the forward and reverse velocity fields, enabled by the momentum field, facilitates velocity field prediction. Additionally, the piecewise linear nature of the trajectory preserves the accelerated denoising capability of Rectified Flow, enabling the generation of high-quality samples with a small number of steps while maintaining high denoising efficiency.

\subsection{Unconditioned Image Generation}
\label{exp:2}

\begin{wraptable}{r}{0.6\textwidth}
\centering
\scriptsize 
\setlength{\tabcolsep}{2pt} 
\renewcommand{\arraystretch}{1.15} 
\captionsetup{font=scriptsize,singlelinecheck=off}
\begin{tabular}{cc|ccc|ccc}  
    \toprule
    \multirow{2}{*}{} & \multirow{2}{*}{} & \multicolumn{3}{c|}{\textbf{CIFAR-10}} & \multicolumn{3}{c}{\textbf{CelebA-HQ}} \\
    \cmidrule(lr){3-5} \cmidrule(lr){6-8}
    $N$& $\hat{Step}$ & FID $\downarrow$ & NFE $\downarrow$ & Recall $\uparrow$ & FID $\downarrow$ & NFE $\downarrow$ & Recall $\uparrow$ \\
    \midrule
    \multirow{3}{*}{1 \textbf{(Rectified Flow)}} 
    & 10  & 36.73 & 10 & 0.391 & 98.98 & 10 & 0.268 \\
    & {50}  & {32.83} & 50 & 
    {0.457} & 65.38 & 50 & 
0.384
 \\
    & 100 & \cellcolor{mycolor_green}{\underline{32.36}} & 100 & \cellcolor{mycolor_green}{\underline{0.473}} & \cellcolor{mycolor_green}{\underline{\textcolor{red!70!black}{58.90}}} & 100 & \cellcolor{mycolor_green}{\underline{\textcolor{red!70!black}{0.445}}} \\
    \midrule
    \multirow{3}{*}{2 \textbf{(Ours)}}
    & 5   & 40.38 & 10 & 0.377 & 84.61 & 10 & 0.345 \\
    & {25}  & {32.55} & 50 & {0.459} & \textcolor{red!70!black}{54.07} & 50 & \textcolor{red!70!black}{0.457} \\
    & 50   & \cellcolor{pearDark!20}{\textbf{32.16}} & 100 & \cellcolor{pearDark!20}{\textbf{0.471}} & \cellcolor{pearDark!20}{\textbf{50.57}} & 100 & \cellcolor{pearDark!20}{\textbf{0.488}} \\
    \midrule
    \multirow{3}{*}{5 \textbf{(Ours)}}
    & 2   & 41.28 & 10 & 0.425 & 110.72 & 10 & 0.177 \\
    & 10  & 41.12 & 50 & 0.454 & 99.57 & 50 & 0.249 \\
    & 20   & 41.91 & 100 & 0.445 & 94.61 & 100 & 0.261 \\
    \bottomrule
\end{tabular}
\caption{Results under different settings on CIFAR-10 and CelebA-HQ datasets. $\hat{Step}$ denotes the number of denoising steps in each sub-path ($\gamma=0.98$).}
\vspace{-1em}
\label{tab:celeba-results}
\end{wraptable}

\textbf{Experiment Settings.}
We build upon the official open-source implementation as the foundation of our model framework, and all experiments are conducted on the image dataset of CIFAR-10~\cite{krizhevsky2009learning} and CelebA-HQ~\cite{karras2018progressive} at a resolution of $256 \times256$. The neural network architecture is adapted with slight modifications from the original design~\cite{liu2023flow}. All models are trained for $70000$ steps with a batch size of $16$ on $8$ NVIDIA A800 GPUs, using the AdamW optimizer~\cite{loshchilov2017fixing} under the default training protocol. To maximize performance within our computational budget, we conduct a grid search over learning rates and weight decay parameters. For evaluation, we generate $30000$ samples from each model and evaluate generating quality and diversity using the Fréchet Inception Distance (FID)~\cite{Seitzer2020FID} and the recall value\cite{sajjadi2018assessing}. As 'recall' is defined as the coverage rate of generated samples over the real data distribution, we evaluate the diversity of generated samples by calculating the Recall value. Additional training and implementation details are provided in Appendix~\ref{appendix:exp-details}.

\textbf{Comparison on CelebA-HQ.}
As shown in Table~\ref{tab:celeba-results}, the momentum flow model consistently achieves superior performance compared to the rectified flow model, as evidenced by significantly lower FID and higher recall values across various settings. While the improvements on CIFAR-10 are relatively modest—possibly due to the simplicity of the dataset, which does not require complex trajectory modeling—the gains on the more challenging CelebA-HQ dataset are significant, achieving an average improvement of over $11$ FID points and $0.06$ recall values. In addition, when reducing the number of function evaluations (NFE) from $100$ to $10$, the performance degradation is minimal, and the model remains competitive with rectified flow under the same sampling budget, highlighting the efficiency of our method grounded in optimal transport (OT). These results indicate that the momentum flow model preserves the fast sampling efficiency of rectified flow while generating higher-quality images. 

\begin{wrapfigure}{r}{0.5\textwidth} 
    \centering
    \includegraphics[width=0.5\textwidth]{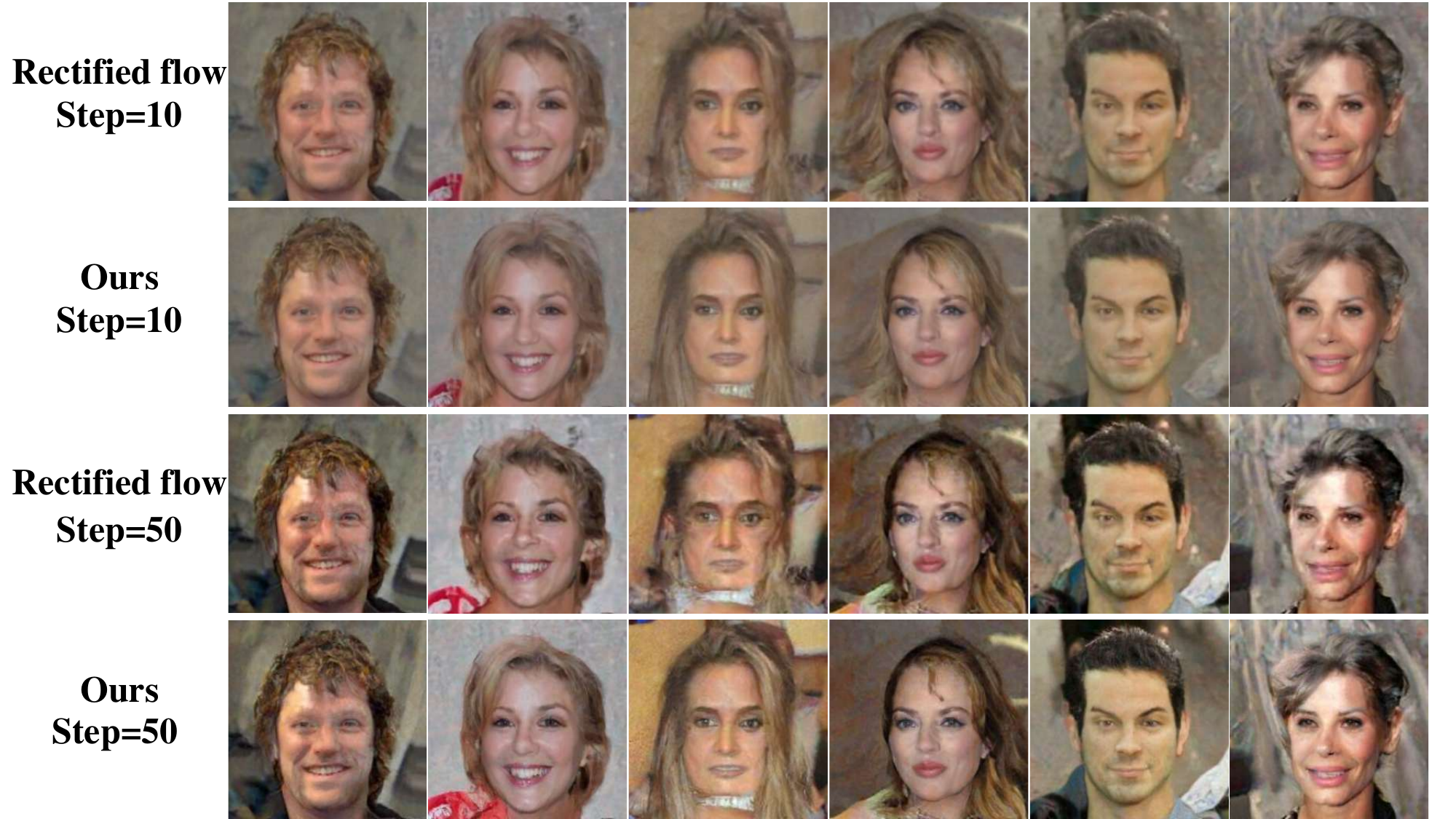}
    \vspace{-8pt}
    \caption{Face generation by Rectified Flow and Momentum Flow under different sampling steps. } 
    \label{fig:ex2}
  \vspace{-10pt}
\end{wrapfigure}

As illustrated in Figure~\ref{fig:ex2}, the deterministic nature of straight-line modeling in rectified flow leads to noticeable distortions in local details (e.g., mouth, eyes, and accessories). In contrast, the momentum flow model employs momentum-guided trajectories to explore a broader space, resulting in significantly improved detail generation. By dynamically injecting controllable velocity deviations via the momentum field, our method enhances both generating diversity and fidelity on high-resolution datasets such as CelebA-HQ, highlighting the effectiveness of multi-scale noise in guiding generation.

\textbf{The Acceleration Process of Momentum Flow.}
We empirically evaluate the efficiency of momentum flow in image generation. Although additional noise is injected into the velocity field, the linear straight structure ensures a constant velocity within each sub-path. This design preserves the efficiency of the original rectified flow. As shown in Figure~\ref{fig:ex2}, we compare the visual quality of generated images under different total denoising step settings: $10$ and $50$ steps. Momentum flow achieves comparable or even superior results to rectified flow while using only half the number of sampling steps, as highlighted in the red-marked values in Table~\ref{tab:celeba-results}. These results demonstrate that momentum flow inherits the acceleration advantages of rectified flows while further benefiting from enhanced flexibility.

\begin{figure}[t]
    \centering
    \includegraphics[width=1\textwidth]{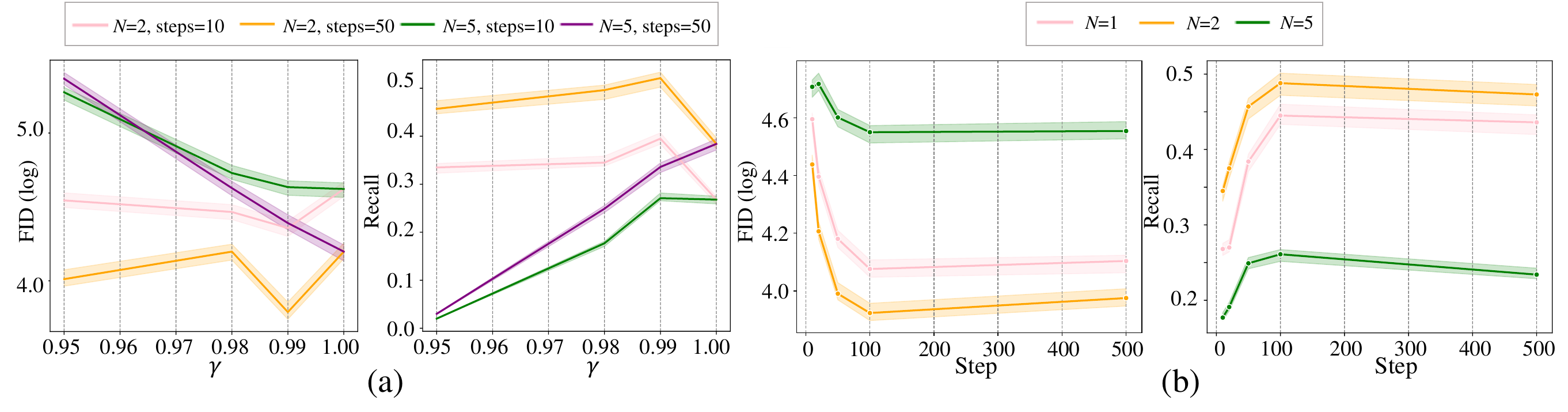}
    \vspace{-16pt}
    \caption{(a) shows the relationship between model performance (FID, Recall) and the $\gamma$ setting. (b) illustrates the relationship between denoising steps and the quality of the generated images.} 
    \label{fig:exgamma}
\vspace{-16pt}
\end{figure}

\textbf{Sampling Efficiency and Generating Diversity.}
Under a fixed image input, we compare the number of sampling steps under the same network architecture to assess their sampling efficiency. The momentum flow model achieves superior performance within the same time budget and requires fewer steps to reach comparable results, demonstrating its strong sampling efficiency. When using the same random seed for generation, the momentum flow model yields higher recall scores than other models, indicating greater diversity in the generated samples. These results confirm that our model effectively balances efficiency and diversity in the generative process, as shown in Figure~\ref{fig:exgamma}.

\textbf{Momentum Decay Coefficient.}
In our model framework, the decay coefficient $\gamma$ controls the level of noise perturbation by modulating the influence of the momentum flow, thereby enabling dynamic refinement of the forward trajectory. We observe that decreasing $\gamma$ causes the momentum flow to deviate more rapidly from the initial momentum direction $\bm{v}_0$, which expands the exploration region and enhances sample diversity. However, excessive decay in the early stages may weaken the guidance from the initial momentum $\bm{v}_0$, so $\gamma$ should not be set too low. As shown in Figure~\ref{fig:exgamma}, when the number of noise-injecting steps is set to $N=2$, a decay coefficient of $\gamma = 0.99$ results in significantly lower FID scores and higher recall values compared to $\gamma = 1$ and $\gamma = 0.98$. In contrast, when $N = 5$, the best performance is achieved with $\gamma = 1$, indicating that a larger number of forward steps $N$ requires a slower decay (i.e., a $\gamma$ closer to $1$) to maintain effective guidance of $\bm v_0$. These experimental results suggest that appropriately selecting the decay coefficient $\gamma$ and the number of velocity steps $N$ can substantially improve both the quality and diversity of generated images.

\textbf{Broader Analysis.}
We compare the sampling processes of rectified flow and momentum flow to assess the advantages of our method in the denoising trajectory. In the early stages of generation—specifically the first few sampling steps—the results of both models appear similar. To illustrate this, we select $\bm z_t$ at $t=0.4$ and perform a single denoising step using the predicted velocity field to obtain a reference image, as shown on the right side of Figure \ref{fig:ex3}. While the early-stage images generated by both methods show no notable differences, particularly in regions such as the hair, momentum flow exhibits a clear advantage in detail fidelity after passing the anchor point ($t=0.5$).

Ignoring the refinement methods like distillation \cite{lee2024improving, zhao2024mobilediffusion}, the initial velocity field often struggles to effectively bridge the gap between the noise and data distributions due to its reliance on fixed straight-line trajectories. The strength of our method lies in its ability to encourage broader exploration of the data space. By introducing momentum-guided velocity deviations, the model is not constrained to a fixed straight-line trajectory. Instead, it gains the flexibility to adjust its path dynamically. 

\begin{figure}[t]
    \centering
    \includegraphics[width=1\textwidth]{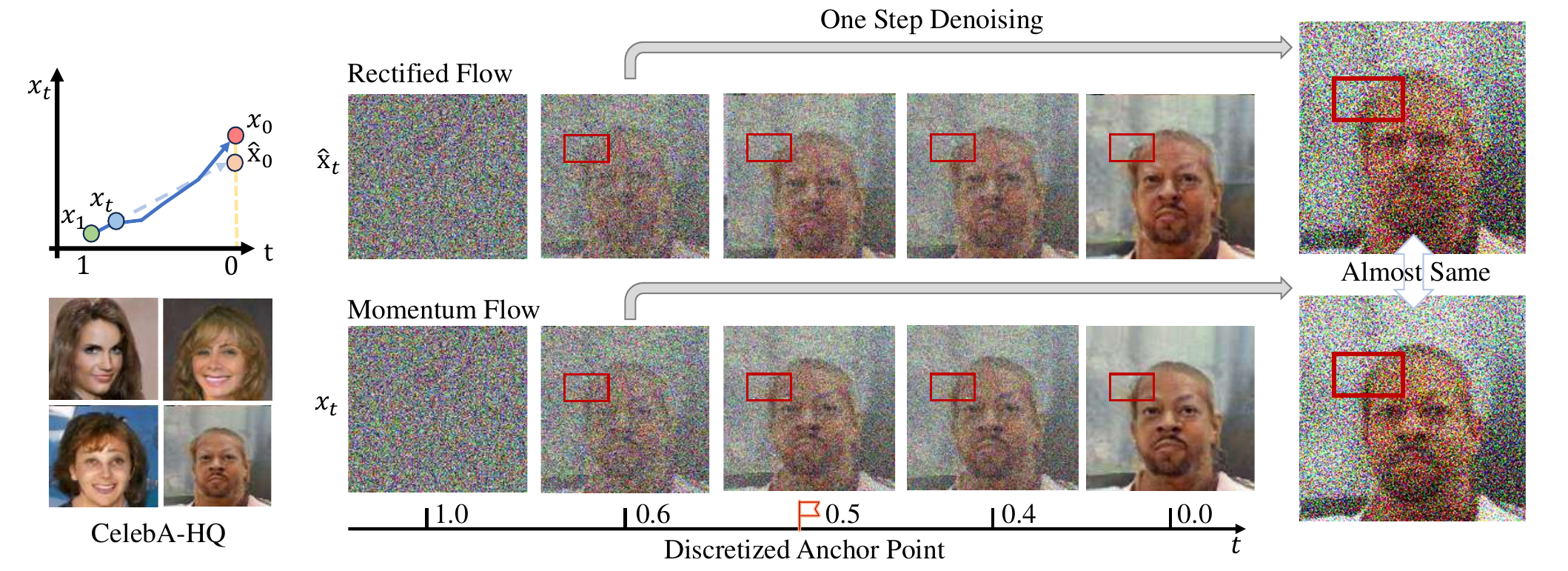}
    \vspace{-16pt}
    \caption{Reverse trajectories of momentum flow and rectified flow at different denoising steps, where the optimized velocity field of momentum flow improves image details.} 
    \label{fig:ex3}
\vspace{-10pt}
\end{figure}

\section{Limitation Discussion}
\label{discussion:limit}
Despite the effectiveness of Discretized-RF in enhancing diversity via momentum-based trajectories, its ability to fully explore diverse sampling paths is partially limited by the choice of the momentum decay coefficient $\gamma_t$. In our current implementation, the decay coefficient $\gamma_t$ remains fixed throughout the forward process, leading to a fixed-rate decay of the initial momentum $\bm{v}_0$. This design choice may hinder the model’s ability to enhance diversity, especially when dealing with complex or multimodal data distributions. We encourage future work to explore adaptive or learnable decay coefficients and to further advance flow matching models by better addressing the optimal transport path problem.

\textbf{Social Impact.}
Our Discretized-RF model enhances both efficiency and diversity in generative models, with promising applications in image enhancement, video generation, and content creation. However, the realism of synthetic images also raises concerns about misuse, such as generating deceptive or harmful content. Responsible deployment, guided by ethical standards and proper safeguards, is essential to mitigate these risks. 

\textbf{More Experiments.}
Due to time and resource constraints, we acknowledge the suboptimal model performance, primarily attributed to opting for a simpler base model and exclusively testing the unconditional scenario. Results from other base models like SD 3.0\cite{esser2024scalingrectifiedflowtransformers} and FLUX 1.0\cite{flux2024} will be included in future updates.
\section{Conclusions}
Discretized-RF proposes an optimal transport method to balance the trade-off between diversity and efficiency. By injecting multi-scale noise perturbations based on momentum flow and formulating discretized straight-line trajectories, our approach effectively optimizes two key limitations of previous models: restricted generating diversity and high computational cost. Extensive experiments show that the momentum flow model achieves both high-quality image generation and fast sampling speed. Looking ahead, we believe the Discretized-RF framework offers a promising direction for designing more flexible flow trajectories and further exploring the diversity-efficiency optimal method.

\begin{ack}
This work is supported by the National Science and Technology Major Project (2023ZD0121403), National Natural Science Foundation of China (No. 62406161), China Postdoctoral Science Foundation (No. 2023M741950), and the Postdoctoral Fellowship Program of CPSF (No. GZB20230347).
\end{ack}
\bibliographystyle{plain}
\bibliography{main,mm_ref}

\begin{thebibliography}{10}

\bibitem{bartosh2024neural}
Grigory Bartosh, Dmitry~P Vetrov, and Christian Andersson~Naesseth.
\newblock Neural flow diffusion models: Learnable forward process for improved diffusion modelling.
\newblock {\em Advances in Neural Information Processing Systems}, 37:73952--73985, 2024.

\bibitem{dalva2024fluxspacedisentangledsemanticediting}
Yusuf Dalva, Kavana Venkatesh, and Pinar Yanardag.
\newblock Fluxspace: Disentangled semantic editing in rectified flow transformers, 2024.

\bibitem{dao2023flowmatchinglatentspace}
Quan Dao, Hao Phung, Binh Nguyen, and Anh Tran.
\newblock Flow matching in latent space, 2023.

\bibitem{esser2024scaling}
Patrick Esser, Sumith Kulal, Andreas Blattmann, Rahim Entezari, Jonas M{\"u}ller, Harry Saini, Yam Levi, Dominik Lorenz, Axel Sauer, Frederic Boesel, Dustin Podell, Tim Dockhorn, Zion English, and Robin Rombach.
\newblock Scaling rectified flow transformers for high-resolution image synthesis.
\newblock In {\em Forty-first International Conference on Machine Learning}, 2024.

\bibitem{esser2024scalingrectifiedflowtransformers}
Patrick Esser, Sumith Kulal, Andreas Blattmann, Rahim Entezari, Jonas Müller, Harry Saini, Yam Levi, Dominik Lorenz, Axel Sauer, Frederic Boesel, Dustin Podell, Tim Dockhorn, Zion English, Kyle Lacey, Alex Goodwin, Yannik Marek, and Robin Rombach.
\newblock Scaling rectified flow transformers for high-resolution image synthesis, 2024.

\bibitem{NEURIPS2024_f0d629a7}
Itai Gat, Tal Remez, Neta Shaul, Felix Kreuk, Ricky T.~Q. Chen, Gabriel Synnaeve, Yossi Adi, and Yaron Lipman.
\newblock Discrete flow matching.
\newblock In A.~Globerson, L.~Mackey, D.~Belgrave, A.~Fan, U.~Paquet, J.~Tomczak, and C.~Zhang, editors, {\em Advances in Neural Information Processing Systems}, volume~37, pages 133345--133385. Curran Associates, Inc., 2024.

\bibitem{NEURIPS2023_d6f764aa}
Martin Gonzalez, Nelson Fernandez~Pinto, Thuy Tran, elies Gherbi, Hatem Hajri, and Nader Masmoudi.
\newblock Seeds: Exponential sde solvers for fast high-quality sampling from diffusion models.
\newblock In A.~Oh, T.~Naumann, A.~Globerson, K.~Saenko, M.~Hardt, and S.~Levine, editors, {\em Advances in Neural Information Processing Systems}, volume~36, pages 68061--68120. Curran Associates, Inc., 2023.

\bibitem{guo2025variationalrectifiedflowmatching}
Pengsheng Guo and Alexander~G. Schwing.
\newblock Variational rectified flow matching, 2025.

\bibitem{ho2020denoising}
Jonathan Ho, Ajay Jain, and Pieter Abbeel.
\newblock Denoising diffusion probabilistic models.
\newblock {\em Advances in neural information processing systems}, 33:6840--6851, 2020.

\bibitem{karras2018progressive}
Tero Karras, Timo Aila, Samuli Laine, and Jaakko Lehtinen.
\newblock Progressive growing of gans for improved quality, stability, and variation.
\newblock In {\em Proceedings of the International Conference on Learning Representations (ICLR)}, 2018.

\bibitem{kawar2022denoising}
Bahjat Kawar, Michael Elad, Stefano Ermon, and Jiaming Song.
\newblock Denoising diffusion restoration models.
\newblock {\em arXiv preprint arXiv:2201.11793}, 2022.

\bibitem{krizhevsky2009learning}
Alex Krizhevsky.
\newblock Learning multiple layers of features from tiny images.
\newblock Technical Report TR-2009, University of Toronto, 2009.

\bibitem{flux2024}
Black~Forest Labs.
\newblock Flux.
\newblock \url{https://github.com/black-forest-labs/flux}, 2024.

\bibitem{lee2024improving}
Sangyun Lee, Zinan Lin, and Giulia Fanti.
\newblock Improving the training of rectified flows.
\newblock {\em Advances in Neural Information Processing Systems}, 37:63082--63109, 2024.

\bibitem{li2024distrifusion}
Muyang Li, Tianle Cai, Jiaxin Cao, Qinsheng Zhang, Han Cai, Junjie Bai, Yangqing Jia, Kai Li, and Song Han.
\newblock Distrifusion: Distributed parallel inference for high-resolution diffusion models.
\newblock In {\em Proceedings of the IEEE/CVF Conference on Computer Vision and Pattern Recognition}, pages 7183--7193, 2024.

\bibitem{NEURIPS2024_9ad996b5}
Senmao Li, Taihang Hu, Joost van~de Weijer, Fahad~Shahbaz Khan, Tao Liu, Linxuan Li, Shiqi Yang, Yaxing Wang, Ming-Ming Cheng, and Jian Yang.
\newblock Faster diffusion: Rethinking the role of the encoder for diffusion model inference.
\newblock In A.~Globerson, L.~Mackey, D.~Belgrave, A.~Fan, U.~Paquet, J.~Tomczak, and C.~Zhang, editors, {\em Advances in Neural Information Processing Systems}, volume~37, pages 85203--85240. Curran Associates, Inc., 2024.

\bibitem{li2024omniflow}
Shufan Li, Konstantinos Kallidromitis, Akash Gokul, Zichun Liao, Yusuke Kato, Kazuki Kozuka, and Aditya Grover.
\newblock Omniflow: Any-to-any generation with multi-modal rectified flows.
\newblock {\em arXiv preprint arXiv:2412.01169}, 2024.

\bibitem{li2023snapfusion}
Yanyu Li, Huan Wang, Qing Jin, Ju~Hu, Pavlo Chemerys, Yun Fu, Yanzhi Wang, Sergey Tulyakov, and Jian Ren.
\newblock Snapfusion: Text-to-image diffusion model on mobile devices within two seconds.
\newblock {\em Advances in Neural Information Processing Systems}, 36:20662--20678, 2023.

\bibitem{lipmanflow}
Yaron Lipman, Ricky~TQ Chen, Heli Ben-Hamu, Maximilian Nickel, and Matthew Le.
\newblock Flow matching for generative modeling.
\newblock In {\em The Eleventh International Conference on Learning Representations}.

\bibitem{liu2022pseudonumericalmethodsdiffusion}
Luping Liu, Yi~Ren, Zhijie Lin, and Zhou Zhao.
\newblock Pseudo numerical methods for diffusion models on manifolds, 2022.

\bibitem{liu2024rfwavemultibandrectifiedflow}
Peng Liu, Dongyang Dai, and Zhiyong Wu.
\newblock Rfwave: Multi-band rectified flow for audio waveform reconstruction, 2024.

\bibitem{liu2022rectifiedflowmarginalpreserving}
Qiang Liu.
\newblock Rectified flow: A marginal preserving approach to optimal transport, 2022.

\bibitem{liu2023flow}
Xingchao Liu, Chengyue Gong, and Qiang Liu.
\newblock Flow straight and fast: Learning to generate and transfer data with rectified flow.
\newblock In {\em The Eleventh International Conference on Learning Representations (ICLR)}, 2023.

\bibitem{liu2023instaflow}
Xingchao Liu, Xiwen Zhang, Jianzhu Ma, Jian Peng, et~al.
\newblock Instaflow: One step is enough for high-quality diffusion-based text-to-image generation.
\newblock In {\em The Twelfth International Conference on Learning Representations}, 2023.

\bibitem{loshchilov2017fixing}
Ilya Loshchilov, Frank Hutter, et~al.
\newblock Fixing weight decay regularization in adam.
\newblock {\em arXiv preprint arXiv:1711.05101}, 5:5, 2017.

\bibitem{luo2024flowdiffuser}
Ao~Luo, Xin Li, Fan Yang, Jiangyu Liu, Haoqiang Fan, and Shuaicheng Liu.
\newblock Flowdiffuser: Advancing optical flow estimation with diffusion models.
\newblock In {\em Proceedings of the IEEE/CVF Conference on Computer Vision and Pattern Recognition}, pages 19167--19176, 2024.

\bibitem{Ma_2024_CVPR}
Xinyin Ma, Gongfan Fang, and Xinchao Wang.
\newblock Deepcache: Accelerating diffusion models for free.
\newblock In {\em Proceedings of the IEEE/CVF Conference on Computer Vision and Pattern Recognition (CVPR)}, pages 15762--15772, June 2024.

\bibitem{ma2024safe}
Zhiyuan Ma, Guoli Jia, Biqing Qi, and Bowen Zhou.
\newblock Safe-sd: Safe and traceable stable diffusion with text prompt trigger for invisible generative watermarking.
\newblock In {\em ACM Multimedia 2024}.

\bibitem{ma2024adapedit}
Zhiyuan Ma, Guoli Jia, and Bowen Zhou.
\newblock Adapedit: Spatio-temporal guided adaptive editing algorithm for text-based continuity-sensitive image editing.
\newblock In {\em Proceedings of the AAAI Conference on Artificial Intelligence}, volume~38, pages 4154--4161, 2024.

\bibitem{ma2024lmd}
Zhiyuan Ma, Zhihuan Yu, Jianjun Li, and Bowen Zhou.
\newblock Lmd: faster image reconstruction with latent masking diffusion.
\newblock In {\em Proceedings of the AAAI Conference on Artificial Intelligence}, volume~38, pages 4145--4153, 2024.

\bibitem{ma2025efficient}
Zhiyuan Ma, Yuzhu Zhang, Guoli Jia, Liangliang Zhao, Yichao Ma, Mingjie Ma, Gaofeng Liu, Kaiyan Zhang, Ning Ding, Jianjun Li, et~al.
\newblock Efficient diffusion models: A comprehensive survey from principles to practices.
\newblock {\em IEEE Transactions on Pattern Analysis and Machine Intelligence}, 2025.

\bibitem{ma2024neural}
Zhiyuan Ma, Liangliang Zhao, Biqing Qi, and Bowen Zhou.
\newblock Neural residual diffusion models for deep scalable vision generation.
\newblock In {\em The Thirty-eighth Annual Conference on Neural Information Processing Systems}, 2024.

\bibitem{Meng_2023_CVPR}
Chenlin Meng, Robin Rombach, Ruiqi Gao, Diederik Kingma, Stefano Ermon, Jonathan Ho, and Tim Salimans.
\newblock On distillation of guided diffusion models.
\newblock In {\em Proceedings of the IEEE/CVF Conference on Computer Vision and Pattern Recognition (CVPR)}, pages 14297--14306, June 2023.

\bibitem{nichol2021improved}
Alexander~Quinn Nichol and Prafulla Dhariwal.
\newblock Improved denoising diffusion probabilistic models.
\newblock In {\em International conference on machine learning}, pages 8162--8171. PMLR, 2021.

\bibitem{ohayon2025posteriormeanrectifiedflowminimum}
Guy Ohayon, Tomer Michaeli, and Michael Elad.
\newblock Posterior-mean rectified flow: Towards minimum mse photo-realistic image restoration, 2025.

\bibitem{sajjadi2018assessing}
Mehdi S.~M. Sajjadi, Olivier Bachem, Mario Lucic, Olivier Bousquet, and Sylvain Gelly.
\newblock Assessing generative models via precision and recall.
\newblock In {\em Advances in Neural Information Processing Systems (NeurIPS)}, 2018.

\bibitem{salimans2022progressivedistillationfastsampling}
Tim Salimans and Jonathan Ho.
\newblock Progressive distillation for fast sampling of diffusion models, 2022.

\bibitem{sauer2024adversarial}
Axel Sauer, Dominik Lorenz, Andreas Blattmann, and Robin Rombach.
\newblock Adversarial diffusion distillation.
\newblock In {\em European Conference on Computer Vision}, pages 87--103. Springer, 2024.

\bibitem{Seitzer2020FID}
Maximilian Seitzer.
\newblock {pytorch-fid: FID Score for PyTorch}.
\newblock \url{https://github.com/mseitzer/pytorch-fid}, August 2020.
\newblock Version 0.3.0.

\bibitem{song2020denoising}
Jiaming Song, Chenlin Meng, and Stefano Ermon.
\newblock Denoising diffusion implicit models.
\newblock {\em arXiv preprint arXiv:2010.02502}, 2020.

\bibitem{song2023consistency}
Yang Song, Prafulla Dhariwal, Mark Chen, and Ilya Sutskever.
\newblock Consistency models.
\newblock 2023.

\bibitem{song2019generative}
Yang Song and Stefano Ermon.
\newblock Generative modeling by estimating gradients of the data distribution.
\newblock {\em Advances in neural information processing systems}, 32, 2019.

\bibitem{song2020score}
Yang Song, Jascha Sohl-Dickstein, Diederik~P Kingma, Abhishek Kumar, Stefano Ermon, and Ben Poole.
\newblock Score-based generative modeling through stochastic differential equations.
\newblock {\em arXiv preprint arXiv:2011.13456}, 2020.

\bibitem{wang2024rectifieddiffusionstraightnessneed}
Fu-Yun Wang, Ling Yang, Zhaoyang Huang, Mengdi Wang, and Hongsheng Li.
\newblock Rectified diffusion: Straightness is not your need in rectified flow, 2024.

\bibitem{wang2024tamingrectifiedflowinversion}
Jiangshan Wang, Junfu Pu, Zhongang Qi, Jiayi Guo, Yue Ma, Nisha Huang, Yuxin Chen, Xiu Li, and Ying Shan.
\newblock Taming rectified flow for inversion and editing, 2024.

\bibitem{wang2024frieren}
Yongqi Wang, Wenxiang Guo, Rongjie Huang, Jiawei Huang, Zehan Wang, Fuming You, Ruiqi Li, and Zhou Zhao.
\newblock Frieren: Efficient video-to-audio generation with rectified flow matching.
\newblock {\em arXiv e-prints}, pages arXiv--2406, 2024.

\bibitem{xu2024ufogen}
Yanwu Xu, Yang Zhao, Zhisheng Xiao, and Tingbo Hou.
\newblock Ufogen: You forward once large scale text-to-image generation via diffusion gans.
\newblock In {\em Proceedings of the IEEE/CVF Conference on Computer Vision and Pattern Recognition}, pages 8196--8206, 2024.

\bibitem{yan2024perflow}
Hanshu Yan, Xingchao Liu, Jiachun Pan, Jun~Hao Liew, Qiang Liu, and Jiashi Feng.
\newblock Perflow: Piecewise rectified flow as universal plug-and-play accelerator.
\newblock {\em arXiv preprint arXiv:2405.07510}, 2024.

\bibitem{Zhang_2024_CVPR}
Huijie Zhang, Yifu Lu, Ismail Alkhouri, Saiprasad Ravishankar, Dogyoon Song, and Qing Qu.
\newblock Improving training efficiency of diffusion models via multi-stage framework and tailored multi-decoder architecture.
\newblock In {\em Proceedings of the IEEE/CVF Conference on Computer Vision and Pattern Recognition (CVPR)}, pages 7372--7381, June 2024.

\bibitem{zhao2024mobilediffusion}
Yang Zhao, Yanwu Xu, Zhisheng Xiao, Haolin Jia, and Tingbo Hou.
\newblock Mobilediffusion: Instant text-to-image generation on mobile devices.
\newblock In {\em European Conference on Computer Vision}, pages 225--242. Springer, 2024.

\bibitem{Zhu_2024_CVPR}
Yixuan Zhu, Wenliang Zhao, Ao~Li, Yansong Tang, Jie Zhou, and Jiwen Lu.
\newblock Flowie: Efficient image enhancement via rectified flow.
\newblock In {\em Proceedings of the IEEE/CVF Conference on Computer Vision and Pattern Recognition (CVPR)}, pages 13--22, June 2024.

\end{thebibliography}


\appendix

\clearpage
\section{Appendix}

\subsection{Proof of the One-step Momentum Update}
\label{appendix:A}
We derive the one-step formula based on the recursive formulation~\eqref{eq:dzdt} of the momentum field $\left\{\bm v_t\right\}_{0}^{T-1}$:
\begin{equation}
\begin{aligned}
\bm v_t & = \sqrt{\gamma_t}\bm v_{t-1} + \sqrt{1-\gamma_t}\beta\bm \epsilon_t\\
&=\sqrt{\gamma_t\gamma_{t-1}}\bm v_{t-2} + \sqrt{\gamma_{t}-\gamma_t\gamma_{t-1}}\beta\bm\epsilon_t+\sqrt{1-\gamma_{t}}\beta\bm \epsilon_{t-1}\\
&=\sqrt{\gamma_t\gamma_{t-1}}\bm v_{t-2}+\sqrt{1-\gamma_t\gamma_{t-1}}\beta\bm\epsilon_t\\
&=\cdots\\
&=\sqrt{\gamma_t\gamma_{t-1}\cdots\gamma_1}\bm v_0 + \sqrt{1-\gamma_t\gamma_{t-1}\cdots\gamma_1}\beta\bm \epsilon_t\\
&=\sqrt{\bar{\gamma_t}}\bm v_0 + \sqrt{1-\bar{\gamma_t}}\beta\bm \epsilon_t
\end{aligned}
\end{equation}
where $\bar{\gamma_t} :=\prod_{i=1}^t \gamma_i$.

\subsection{Proof of the One-step Forward Data Distribution}
\label{appendix:B}
In the forward process, the data distribution evolves under the momentum field $\left\{\bm v_t\right\}_{0}^{T-1}$. According to the one-step momentum update formula~\eqref{eq:3}, the forward data distribution $\bm\pi(\bm z_t)$ originates from the $\bm{\pi}_0$, perturbed by a exponentially scaled decaying contribution of the initial momentum $\bm{v}_0$, along with random noise $\bm{\epsilon_t}$,
\begin{equation}
\begin{aligned}
\bm z_t & =\bm z_{t-1}+\bm v_{t-1} \\
& =\bm z_0+\bm v_0+\bm v_1+\bm v_2+\cdots+\bm v_{t-1} \\
& =\bm z_0+\bm v_0(1+\sqrt{\gamma}+\sqrt{\gamma ^2}+\cdots+\sqrt{\gamma^{t-1}})+\sum_{i=1}^{t-1} \sqrt{1-\gamma^i} \beta\bm \epsilon_i  \\
& =\bm z_0+\bm v_0(\frac{\sqrt{\gamma^t}-1}{\sqrt{\gamma}-1})+\sqrt{t-(1+\gamma+\gamma^2+\cdots+\gamma^{t-1})}\beta\bm \epsilon_t  \\
& =\bm z_0+(\bm \epsilon_0-\bm z_0)(\frac{\sqrt{\gamma^t}-1}{\sqrt{\gamma}-1}) \beta+\sqrt{t-\frac{\gamma^t-1}{\gamma-1}}\beta\bm \epsilon_t  \\
& =(1-(\frac{\sqrt{\gamma^t}-1}{\sqrt{\gamma}-1}) \beta) \bm z_0+\sqrt{(\frac{\sqrt{\gamma^t}-1}{\sqrt{\gamma}-1})^2-\frac{\gamma^t-1}{\gamma-1}+t}\beta\bm \epsilon_t.
\end{aligned}
\end{equation}
Thus we have
\begin{equation}
    q(\bm z_t \vert \bm z_{0}) = \mathcal{N}\left(\bm z_t; (1-(\frac{\sqrt{\gamma^t}-1}{\sqrt{\gamma}-1}) \beta) \bm z_{0}, ((\frac{\sqrt{\gamma^t}-1}{\sqrt{\gamma}-1})^2-\frac{\gamma^t-1}{\gamma-1}+t)\beta^2\mathbf{I}\right).
\end{equation}

\subsection{Experiment Details}

\textbf{Dataset Description:} We use three datasets for training:  
\begin{enumerate}
    \item CIFAR-10: Images with a resolution of $32 \times 32$ from the CIFAR-10 training set.
    \item CelebA-HQ: Images from the `img\_align\_celeba\_png.7z` version of the CelebA-HQ dataset, resized to $256 \times 256$.
    \item During training, images are normalized to have zero mean and unit variance.
    \item ImageNet: Images from ImageNet resized to $32 \times 32$ and $64 \times 64$.
\end{enumerate}
During training, images are normalized to have zero mean and unit variance.

\textbf{Training Details:} The model is trained using the Adam optimizer with a learning rate of $3 \times 10^{-4}$. For ImageNet, we use a batch size of $64$, whereas for other datasets, we use a batch size of $16$. Training is conducted for $70000$ steps. A cosine annealing learning rate scheduler is used.

\textbf{Implementation Details:} The experiments are implemented in PyTorch (version 2.6.0) and conducted on an NVIDIA A800-SXM4-80GB GPU. Random seeds are set to $42$ for reproducibility.

\textbf{Performance Details:} Given that our training commenced from scratch and employed a relatively simple network architecture (U-net), our baseline performance is not as robust as that of most diffusion models with more intricate designs. However, since all our experiments were conducted within the same architectural framework that we designed, our comparisons remain fair and persuasive.

\textbf{Additional Experiment Settings:} Ablation studies are conducted to analyze the impact of different hyperparameters. Hyperparameter tuning is performed using grid search over the learning rate and batch size.

\subsection{Experiments on ImageNet}
\begin{figure}[t]
    \centering
    \includegraphics[width=0.8\textwidth]{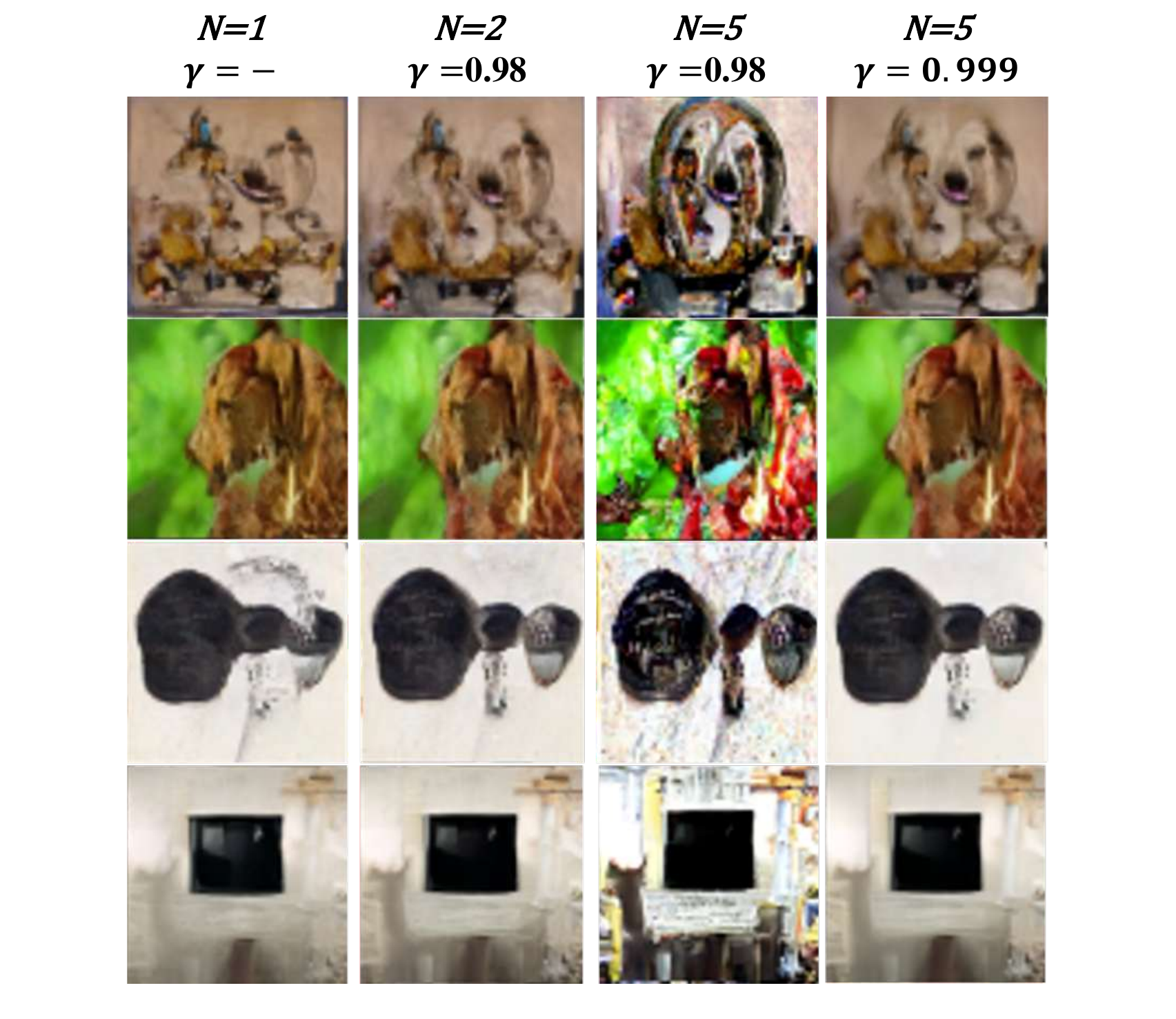}
    \vspace{-10pt}
    \caption{The impact of adjusting $N$ and $\gamma$ on image details in Momentum Flow ($Step=50$). '-' means $\gamma$ do not influence the trajectories while $N=1$.} 
    \label{fig:ex5}
\vspace{-10pt}
\end{figure}

\begin{table}
\centering
\scriptsize 
\setlength{\tabcolsep}{2pt} 
\renewcommand{\arraystretch}{1.15} 
\captionsetup{font=scriptsize,singlelinecheck=off}
\begin{tabular}{cc|ccc|ccc}  
    \toprule
    \multirow{2}{*}{} & \multirow{2}{*}{} & \multicolumn{3}{c|}{\textbf{ImageNet-32}} & \multicolumn{3}{c}{\textbf{ImageNet-64}} \\
    \cmidrule(lr){3-5} \cmidrule(lr){6-8}
    $N$& $\hat{Step}$ & FID $\downarrow$ & NFE $\downarrow$ & Recall $\uparrow$ & FID $\downarrow$ & NFE $\downarrow$ & Recall $\uparrow$ \\
    \midrule
    \multirow{3}{*}{1 \textbf{(Rectified Flow)}} 
    & 10  & 42.864& 10 & 0.319& 61.48& 10 & 0.366\\
    & {50}  & 28.001& 50 & 
    0.395& 42.42& 50 & 
0.457\\
    & 100 & \textbf{25.82}& 100 & 0.398& 41.83& 100 & 0.451\\
    \midrule
    \multirow{3}{*}{2 \textbf{(Ours $\gamma=0.98$)}}
    & 5   & 41.56& 10 & 0.334& 60.34& 10 & 0.368\\
    & {25}  & 28.03& 50 & \textbf{0.403}& 41.99& 50 & 0.454\\
    & 50   & 26.15& 100 & 0.400& 41.77& 100 & 0.459\\
    \midrule
    \multirow{3}{*}{5 \textbf{(Ours $\gamma=0.98$)}}
    & 2   & 50.06& 10 & 0.343& 104.06& 10 & 0.258\\
    & 10  & 49.85& 50 & 0.342& 96.70& 50 & 0.294\\
    & 20   & 49.13& 100 & 0.349& 94.87& 100 & 0.294\\
    \midrule
    \multirow{3}{*}{5 \textbf{(Ours $\gamma=0.999$)}}
    & 2   &45.12 & 10 &0.329 & 59.58& 10 &0.381 \\
    & 10  &30.74 & 50 &0.398 &\textbf{42.20} & 50 &\textbf{0.476} \\
    & 20   &28.86 & 100 &0.390 & 41.71& 100 &0.461 \\
    \bottomrule
\end{tabular}
\caption{Results under different settings on ImageNet datasets. $\hat{Step}$ denotes the number of denoising steps in each sub-path.}
\vspace{-1em}
\label{tab:imagenet-results}
\end{table}
As shown in Table~\ref{tab:imagenet-results}, the momentum flow model consistently achieves significant improvements on ImageNet. By balancing  $N$ and $\gamma$,we verify that more anchor points bring greater gains and $\gamma$ should increase with the number of anchor points to reduce velocity abruptness in momentum field.As shown in \ref{fig:ex5}, a larger $N$ brings more significant diversity, but it may compromise image quality, so the noise intensity $\gamma$ should be correspondingly adjusted. The momentum flow model retains the fast sampling efficiency of rectified flow, while also generating higher-quality images, as shown by these results.

\begin{figure}[t]
    \centering
    \includegraphics[width=1\textwidth]{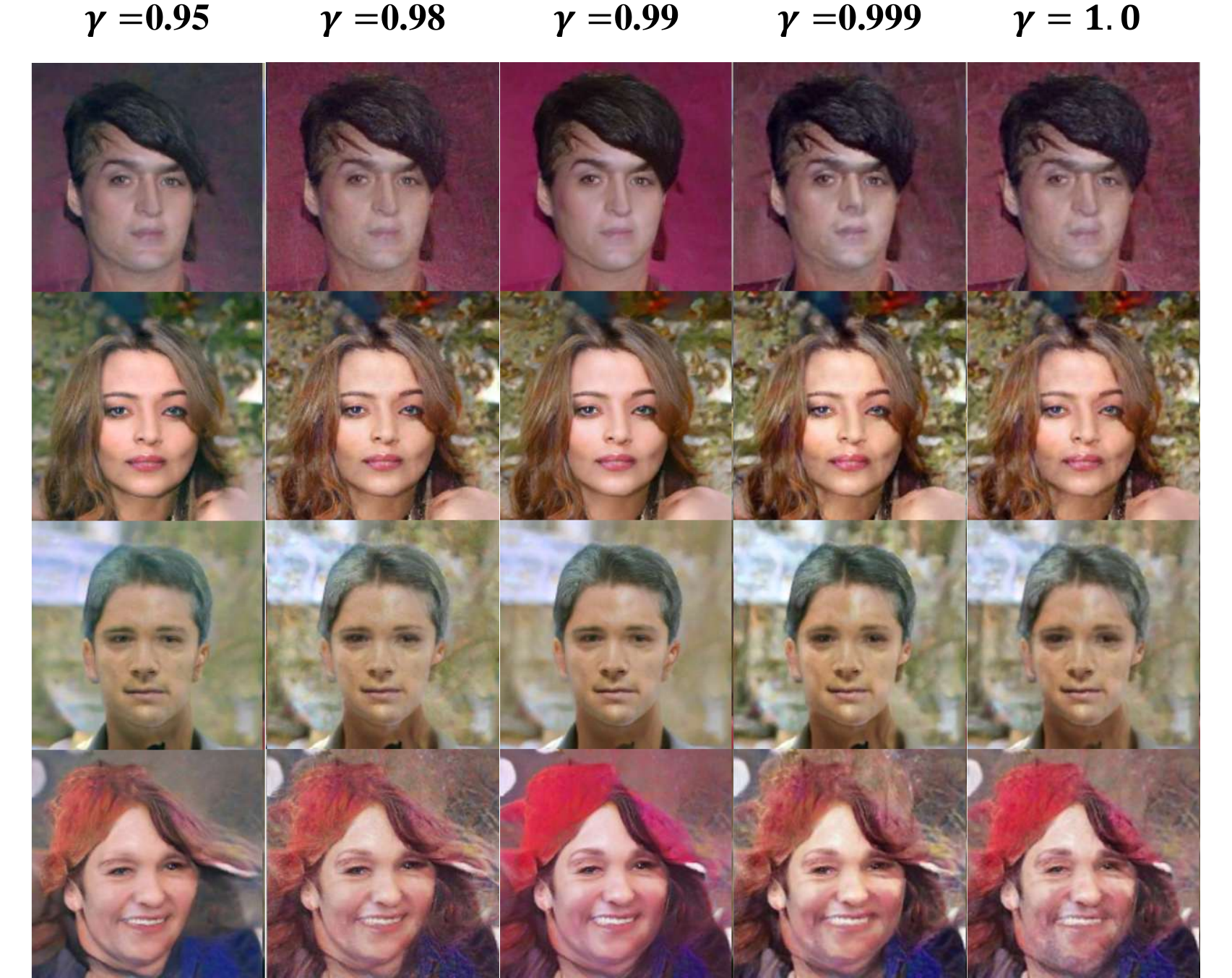}
    \vspace{-16pt}
    \caption{Generated images under different values of gamma, where both excessively large and excessively small gamma values can deteriorate image quality ($N=2, Step=50$).} 
    \label{fig:ex3}
\vspace{-10pt}
\end{figure}
\label{appendix:exp-details}


\clearpage
\section*{NeurIPS Paper Checklist}

\end{document}